\documentclass[journal, onecolumn, 11 pt]{IEEEtran}
\newlength{\tocsep}
\usepackage{makeidx}
\usepackage{epsfig,float,multirow,footmisc,rotating}
\usepackage{graphics}
\usepackage{graphicx}
\usepackage{color}
\usepackage{placeins}

\usepackage[utf8]{inputenc}
\usepackage{verbatim}
\usepackage{url}
\usepackage{array}
\DeclareGraphicsExtensions{.jpg, .png, .pdf, .tiff}

\usepackage{wrapfig}
\usepackage{sidecap}
\usepackage{multicol}

\usepackage[labelfont=bf]{caption}
\usepackage{subcaption}
\usepackage{multimedia}
\usepackage[]{units}
\usepackage{amsmath,amsfonts,amssymb}
\usepackage[normalem]{ulem}
\usepackage{environ}
\usepackage{afterpage}

\definecolor{darkgreen}{rgb}{0.0, 0.2, 0.13}
\definecolor{darkolivegreen}{rgb}{0.33, 0.42, 0.18}

\newcommand{\eg}{\textit{e.g.}\ }
\newcommand{\ie}{\textit{i.e.}\ }

\newcommand{\etal}{\textit{et al.}\ }

\begin{document}

\title{
How to Blend a Robot within a Group of Zebrafish: Achieving Social Acceptance through Real-time Calibration of a Multi-level Behavioural Model
}

\author{
Leo Cazenille$^{1,2}$,
Yohann Chemtob$^{1}$,
Frank Bonnet$^{3}$,
Alexey Gribovskiy$^{3}$,\\
Francesco Mondada$^{3}$,
Nicolas Bredeche$^{2}$,
Jos\'{e} Halloy$^{1}$\\

{1} Univ Paris Diderot, Sorbonne Paris Cit\'e, LIED, UMR 8236, 75013, Paris, France
\\{2} Sorbonne Universit\'e, CNRS, ISIR, F-75005 Paris, France
\\{3} Robotic Systems Laboratory, School of Engineering, Ecole Polytechnique F\'ed\'erale de Lausanne, ME B3 30, Station 9, 1015 Lausanne, Switzerland
}

\maketitle

\begin{abstract}
We have previously shown how to socially integrate a fish robot into a group of zebrafish thanks to biomimetic behavioural models. The models have to be calibrated on experimental data to present correct behavioural features. This calibration is essential to enhance the social integration of the robot into the group. When calibrated, the behavioural model of fish behaviour is implemented to drive a robot with closed-loop control of social interactions into a group of zebrafish. This approach can be useful to form mixed-groups, and study animal individual and collective behaviour by using biomimetic autonomous robots capable of responding to the animals in long-standing experiments. Here, we show a methodology for continuous real-time calibration and refinement of multi-level behavioural model. The real-time calibration, by an evolutionary algorithm, is based on simulation of the model to correspond to the observed fish behaviour in real-time. The calibrated model is updated on the robot and tested during the experiments. This method allows to cope with changes of dynamics in fish behaviour. Moreover, each fish presents individual behavioural differences. Thus, each trial is done with naive fish groups that display behavioural variability. This real-time calibration methodology can optimise the robot behaviours during the experiments. Our implementation of this methodology runs on three different computers that perform individual tracking, data-analysis, multi-objective evolutionary algorithms, simulation of the fish robot and adaptation of the robot behavioural models, all in real-time.
\end{abstract}


\section{Introduction}
The study of animal collective behaviour involves the search for the  relevant signals and mechanisms used by the animals for social interactions~\cite{patricelli2010robotics,knight2005animal}.
Robots can help ethologists to test various hypothesis on the nature of these signals by inducing specific and controlled stimuli to assess animal response. 

Autonomous robots are capable to interact with animals and can serve as tools to study social dynamics~\cite{mondada2013general}.
This approach has already been used in studies to analyse the behaviour of ducks~\cite{vaughan2000experiments}, drosophila~\cite{zabala2012simple}, cockroaches~\cite{halloy2007social}, fish~\cite{cazenille2017acceptation,landgraf2014blending,landgraf2016robofish,kim2018closed,Katzschmanneaar3449,bonnet2018closed,bierbach2018using}, bees~\cite{griparic2017robotic,landgraf2012imitation,stefanec2017governing} and birds~\cite{jolly2016animal,de2011influence,gribovskiy2018designing}.

Here, we socially integrate a behavioural biomimetic robotic lure into a group of four zebrafish (\textit{Danio rerio}) moving in a structured environment and validate its acceptance by the animals.
This problem is difficult because the robotic lure must be designed to be perceived as a social companion by the animals: it must, to a certain extent look like a fish, behave like a fish, be able to respond appropriately to environmental and social cues to close the loop of social interactions with the fish. Closing the loop of social interactions requires real-time individual perception and a decision-making algorithm to control the robot behaviours~\cite{cazenille2017acceptation}. 

These aspects were investigated in~\cite{cazenille2017acceptation,cazenille2017automated} through the use of biomimetic robotic fish lures driven by a calibrated biomimetic model to make the robot mimics expected fish behaviour. 
An evolutionary algorithm (NSGA-II~\cite{deb2002fast}) was used to optimise the parameters of this model so that the resulting collective dynamics corresponded to those observed in biological experiments.
This type of controller allowed the robot to be a real group-member making its own decisions rather than a passive follower. 

However, the model calibration was done off-line and not during the ongoing experiments. As such, it could not take into account the changes in animal behaviour across experiments and the intrinsic behavioural differences between groups used in experiments.

Here, we tackle this problem by continuously refining and calibrating the biomimetic model driving the robot behaviours in real-time during the experiment by using on-line evolutionary algorithm (NSGA-II~\cite{deb2002fast}).
This task is computationally-intensive and requires three computers to deal with agent real-time tracking, robot control, real-time data-analysis, and model calibration.
We test this methodology in a set of 10 experiments with four fish and one robot. In each case, the robot closed-loop behaviour becomes progressively socially integrated into the group of fish. This is the first step towards evolving mixed-group of animals and robot~\cite{cazenille2017acceptation}.

\afterpage{
\begin{figure}[H]
\begin{center}
\includegraphics[width=0.99\textwidth]{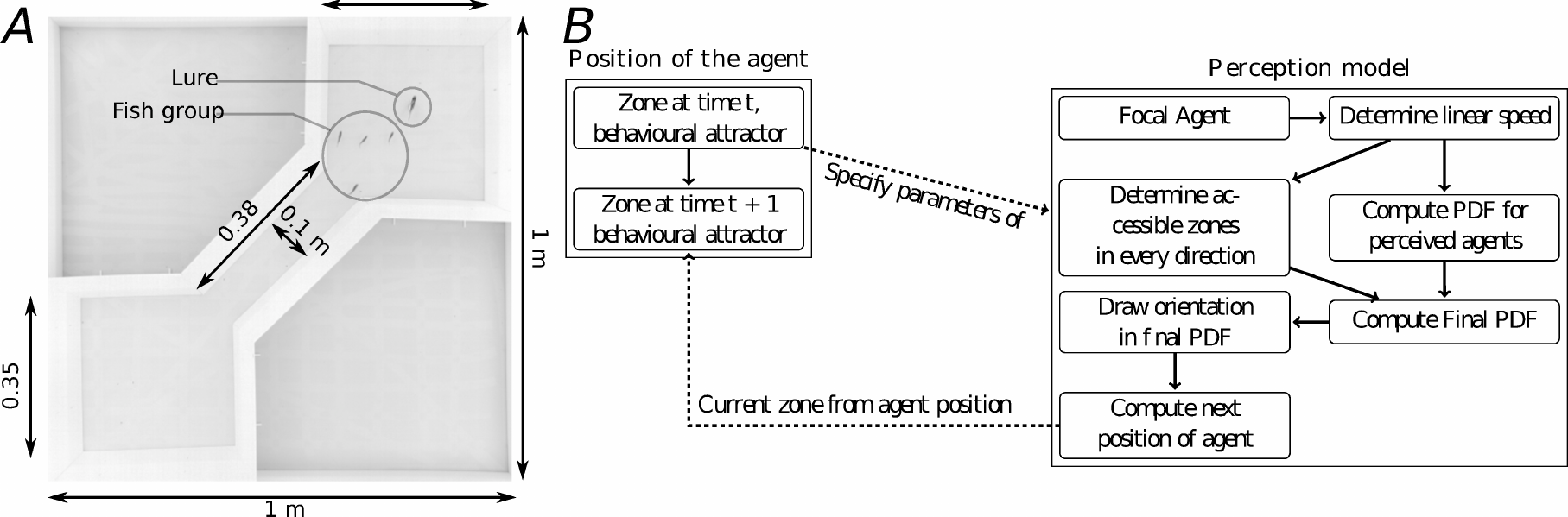}
\caption{ \textbf{A.} Experimental setup: a tank with two square rooms ($350\times350$~mm at floor level) connected by a corridor ($380\times100$~mm). This set-up is used to study zebrafish collective behaviours~\cite{collignon2016stochastic,seguret2016strain,seguret2017loose,cazenille2017acceptation,cazenille2017automated}. It is composed of three zones (corridor, center of the rooms, close to room walls) that correspond to three different behavioural attractors.
\textbf{B.} Multilevel model for fish behaviour~\cite{cazenille2017acceptation,cazenille2017automated}. The agents behave differently depending on the zone where they are situated. }
\label{fig:setup}
\end{center}
\end{figure}
}

\section{Materials and Methods} \label{sec:methods}

\subsection{Experimental set-up} \label{sec:setup}
We use the experimental set-up from~\cite{cazenille2017automated,bonnet2017cats,seguret2017loose,cazenille2017acceptation,collignon2017collective} (Fig.~\ref{fig:workflow}, "Control \& tracking" part) with a white plexiglass arena~(Fig.~\ref{fig:setup}A) of $1000\times1000\times100$~mm composed of two rooms linked by a corridor.
We use the FishBot robot~\cite{bonnet2014miniature,Bonnet2016IJARS,bonnet2017cats}, powered by two conductive plates under the aquarium, to interact with fish. An overhead camera captures frames (15 FPS, $500\times500$px), that are then tracked to find the fish positions. A complementary fish-eye camera (15 FPS, $640\times480$px) placed under the fish tank is used to track the position of the robot.

We used 10 groups of 4 adults wild-type AB zebrafish (\textit{Danio rerio}) in ten 30-minutes experiments as in~\cite{cazenille2017acceptation,cazenille2017automated,seguret2017loose}: 30 minutes is sufficient to capture the behaviour and dynamics of groups of 4 zebrafish.
Fish are released in the aquarium after the lure is placed in the aquarium.

To ensure real-time adaptation, our methodology is computationally intensive, and uses three networked 32-core computers (Fig.~\ref{fig:workflow}). Computer 1 is used to track the agents in real-time and control the robot according to the behavioural model of Sec.~\ref{sec:model}. Computer 2 performs every $60s$ data-analysis on the tracked positions of agents from Computer 1, and estimates the biomimetism of robot behaviour (which, in our case, can be viewed as a metric of social integration as defined in~\cite{cazenille2017acceptation}). Computer 3 re-calibrates every $60s$ the behavioural model to correspond as close as possible to the behaviour of experimental fish (measured by Computer 2). The resulting calibrated parameter set is then sent to Computer 1 to serve as parameters of the robot controller model. It allows the robot to progressively mimics the behaviour of the fish and be socially accepted.

The experiments performed in this study were conducted under the authorisation of the Buffon Ethical Committee (registered to the French National Ethical Committee for Animal Experiments \#40) after submission to the French state ethical board for animal experiments.

\subsection{Behavioural model} \label{sec:model}
We use the multi-level model from~\cite{cazenille2017automated,cazenille2017acceptation} (inspired from~\cite{collignon2016stochastic}) that describes the individual and collective behaviours of fish (Fig.~\ref{fig:setup}B). This model takes into account both social interactions and environmental cues (\ie walls and structure of the tanks). It is stochastic, multi-level and context-dependent.

Fish behave differently depending on their spatial position. Namely, this model identify three zones of the structured set-up with different fish behaviours (Fig.~\ref{fig:setup}A): when they are close to the walls, when they are in the centre of the rooms, and when they pass through the corridor. Near the walls, fish perform mainly thigmotactism (wall following) while in room centre they exhibit exploratory behaviour. In the corridor, they tend to go in a straight line with increased speed to reach the subsequent room. Fish also react to social cues leading to collective behaviour such as collective departures from the rooms~\cite{collignon2017collective}.
Very few models of fish collective behaviours take into account the presence of walls~\cite{collignon2016stochastic,calovi2018disentangling}.

The agents update their position vector ${X_i}$ with a velocity vector ${V_i}$:
\begin{equation}
X_i(t+\delta t)= X_i(t) + V_i(t)\delta t
\label{equa:zebrapos}
\end{equation}
\begin{equation}
V_i(t+\delta t)= v_i (t+\delta t) \Theta_i (t+\delta t)
\label{equa:zebraspeed}
\end{equation}
with $v_i$ the linear speed of the $i^{th}$ agent and $\Theta_i$ its orientation. The linear speed $v_i$ of the agent is randomly drawn from the experimentally measured instantaneous speed distribution.

The orientation $\Theta_i$ is drawn from probability density function (PDF) computed as a mixture distribution of von Mises distributions centred on the stimuli perceived by the focal agent.
It takes into account the influence of other agents and of the walls of the experimental arena. The resulting PDF is composed of the weighted sum of (i) a PDF taking into account the effect of the walls and (ii) a PDF describing the response to other agents. The parameter $\gamma_{z_1, z_2}$, used as a multiplicative term of the final PDF, modulates the attraction of agents towards target zones.

We numerically compute the cumulative distribution function (CDF) corresponding to this final PDF by performing a cumulative trapezoidal numerical integration of the PDF in the interval $[-\pi,\pi]$. Then, the model draws a random direction $\Theta_i$ in this distribution by inverse transform sampling. The position of the fish is then updated according to this direction and his velocity.

\afterpage{
\begin{figure}[H]
\begin{center}
\includegraphics[width=0.99\textwidth]{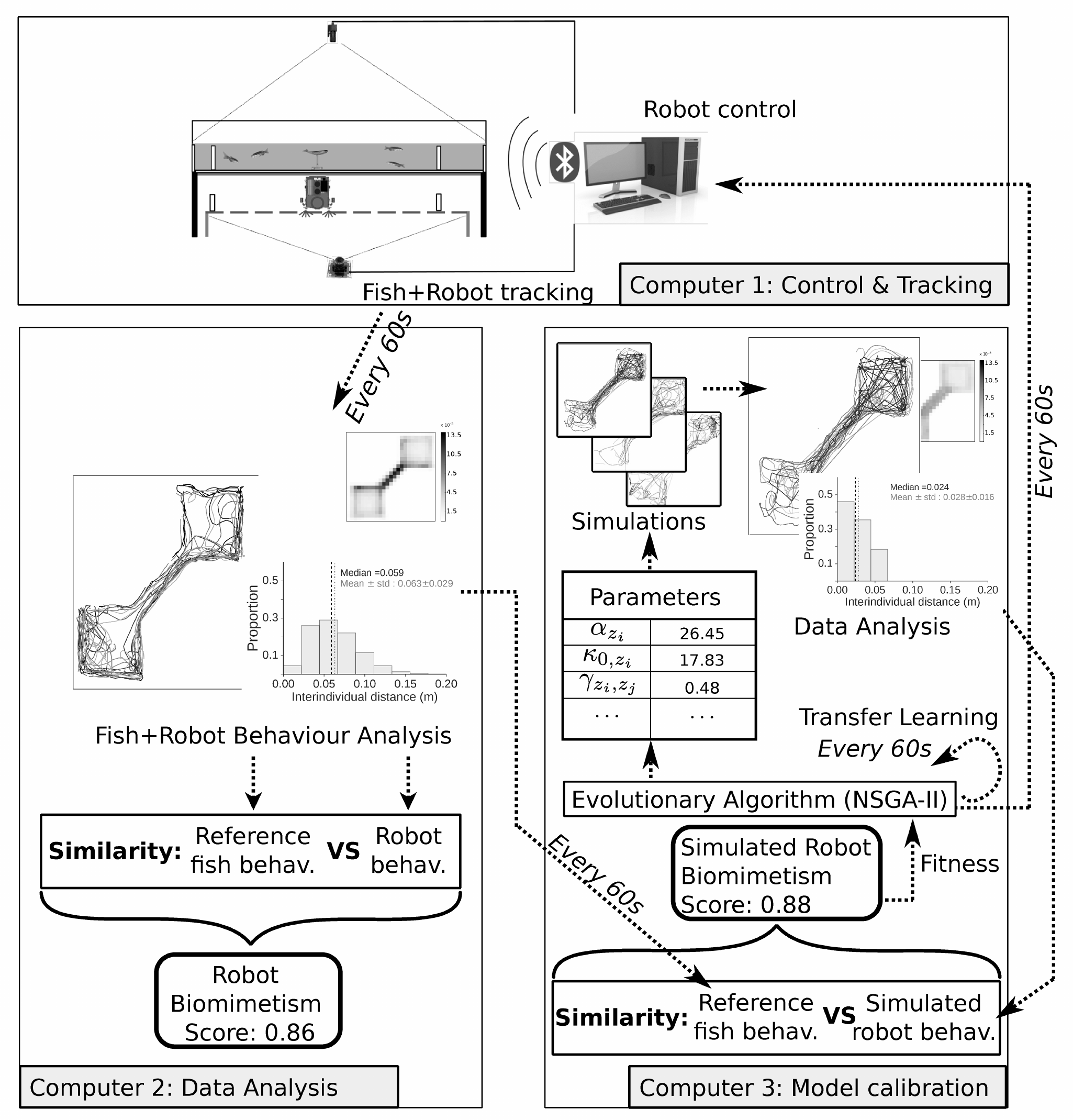}
\caption{Workflow of our real-time calibration methodology. It involves extensive computation to be able to function in real-time, and thus is implemented over three 32-core computers. Computer 1 tracks the positions of fish and robot and is also running the robot controller. Computer 2 performs data-analysis of the fish and the robot behaviour using the data gathered by Computer 1 during 60s. Computer 3 calibrates the behavioural model (presented in Sec.~\ref{sec:model}) to be as close as possible to the observed behaviour of the fish as assessed by the data-analysis performed by Computer 2. It also uses the knowledge acquired during the previous calibration processes. The calibrated model is sent every 60s to Computer 1 to be used to drive the robot. The social acceptation of the robot behaviour is measured by Computer 2 with a distance metric of collective features.}
\label{fig:workflow}
\end{center}
\end{figure}
\clearpage
}

\subsection{Communication between computer nodes} \label{sec:communication}
We connect the three computers (Fig.~\ref{fig:workflow}) using the ZeroMQ distributed messaging protocol~\cite{hintjens2013zeromq}: computers receiving messages act as ZeroMQ subscribers, and computers sending messages act as ZeroMQ publishers.

The tracked agent trajectories are compiled in the form of trajectory files, and sent every $60s$ from Computer 1 to Computer 2 through the \verb!rsync!~\cite{rsync} command line application (a process which usually only need $2s$ to $3s$ that is sufficient because the parameter update is every $60s$). Then, Computer 1 send a ZeroMQ message to Computer 2 to acknowledge that the transfer is completed. Data-analysis scores from Computer 2 to Computer 3, and model parameters from Computer 3 to Computer 1 are sent every $60s$ through ZeroMQ messages.

\subsection{Real-time tracking} \label{sec:trackingAndControl}
We use the CATS framework~\cite{bonnet2017cats} to track agents (fish and robot) in real-time, on Computer 1. Fish are tracked (but not identified) by using frames captured by the overhead camera (Fig.~\ref{fig:workflow}) through the Shi-Tomasi method \cite{Shi1994} implemented in the OpenCV library~\cite{opencv_library}. In parallel, the robot is tracked through the video frames from the camera below the fish-tank by colour and contours detection~\cite{Suzuki1985}.
Every $60s$ the positions of the agents are sent to Computer 2 for data-analysis.

\subsection{Data-analysis} \label{sec:dataAnalysis}
Every $60s$, Computer 2 calculates the behavioural statistics using the tracked positions of agents (from computer 1) over the last $120s$ of the running experiment, for all three zones of the arena. For a zone $e$, these statistics are: the distribution of inter-individual distances between agents ($D_{e}$), the distribution of distances of agents to their nearest wall ($W_{e}$), the distribution of zones occupation ($O_{e}$), the transition probabilities from zone $e$ to others ($T{e}$).

These statistics are computed either only on fish agents (\textbf{Control} case: $e_c$) or on fish and robotic agents (\textbf{Robot Social Integration} case: $e_r$).
We define a similarity score (ranging from $0.0$ to $1.0$) to measure the biomimetism of robot behaviour compared to the \textbf{Control} case:
\begin{equation}
S(e_r, e_c) = \sqrt[4]{I(D_{e_r}, D_{e_c}) I(W_{e_r}, W_{e_c}) I(O_{e_r}, O_{e_c}) I(T_{e_r}, T_{e_c})}
\end{equation}
The function $I(P, Q)$ is defined as such: $I(P, Q) = 1 - H(P, Q)$.
The $H(P, Q)$ function is the Hellinger distance between two histograms ~\cite{deza2006dictionary}. It is defined as: $H(P, Q) = \frac{1}{\sqrt{2}} \sqrt{ \sum_{i=1}^{d} (\sqrt{P_i} - \sqrt{Q_i}  )^2 }$ where $P_i$ and $Q_i$ are the bin frequencies.

Cazenille~\etal~\cite{cazenille2017acceptation,cazenille2017automated} demonstrated that robotic lures with biomimetic morphology and behaviour are be more socially integrated into the group of fish than non-biomimetic lures. As such, the biomimetism score defined earlier corresponds to the social acceptatation of the robot by the fish.

When this statistics and scores are computed, they are dispatched to Computer 3 (by the ZeroMQ system described in Sec.~\ref{sec:communication}) to guide the optimisation process.

\subsection{Real-time optimisation of model parameters} \label{sec:optim}
We design a calibration methodology (Fig.~\ref{fig:workflow}) capable of optimising in real-time the parameters of the behavioural model from Sec.~\ref{sec:model} to mimic as close as possible to the behaviour of experimental fish. The behavioural similarity is quantified as described in Sec.~\ref{sec:dataAnalysis}.

It is inspired from the off-line calibration methodology in~\cite{cazenille2017automated} and uses the NSGA-II~\cite{deb2002fast} multi-objective global optimiser (population of 60 individuals, 300 generations) with three objectives to maximise.
We define a fitness with three objectives: the first objective is a performance objective corresponding to the $S_(e_1, e_2)$ function. Two other objectives are considered to guide the evolutionary process: one that promotes genotypic diversity~\cite{mouret2012encouraging} (defined by the mean euclidean distance of the genome of an individual to the genomes of the other individuals of the current population), the other encouraging behavioural diversity (defined by the euclidean distance between the $D_{e}$, $W_{e}$, $O_{e}$ and $T_{e}$ scores of an individual).

This process is performed and restarted every $60s$ on Computer 3 (starting $120s$ after the beginning of the experiment to gather data, Fig.~\ref{fig:workflow}) using data gathered during the last $120s$. Every restart of the evolutionary algorithm keeps the last generation of individuals evolved during the previous round of evolution to bootstrap the current round of evolution, a system akin to transfer learning. On our 32-core computer, one generation is computed approximately every $4s$, so around $15$ generations are computed at every evolutionary round.

We do not optimise the linear speed $v_i$ of the agents. It is randomly drawn from the experimental speed distribution. We use the NSGA-II implementation provided by the DEAP python library~\cite{fortin2012deap}.

\subsection{Robot implementation and control} \label{sec:robot}
The robot is driven by the model presented in Sec.~\ref{sec:model} thanks to the CATS framework~\cite{bonnet2017cats}. The model is calibrated every $60s$ using the methodology of Sec.~\ref{sec:optim}, in experiments involving four fish and one robot. Every $200ms$, the tracked positions of the four fish are integrated into the model to compute the target position of a fifth agent. The robot is programmed to follow this target position by using the biomimetic movement patterns as in~\cite{Bonnet2016IJARS,cazenille2017acceptation}.

\section{Results} \label{sec:results}
\afterpage{

\clearpage
\begin{figure}[H]
\begin{center}
\begin{subfigure}[b]{0.45\textwidth}
\includegraphics[width=0.99\textwidth]{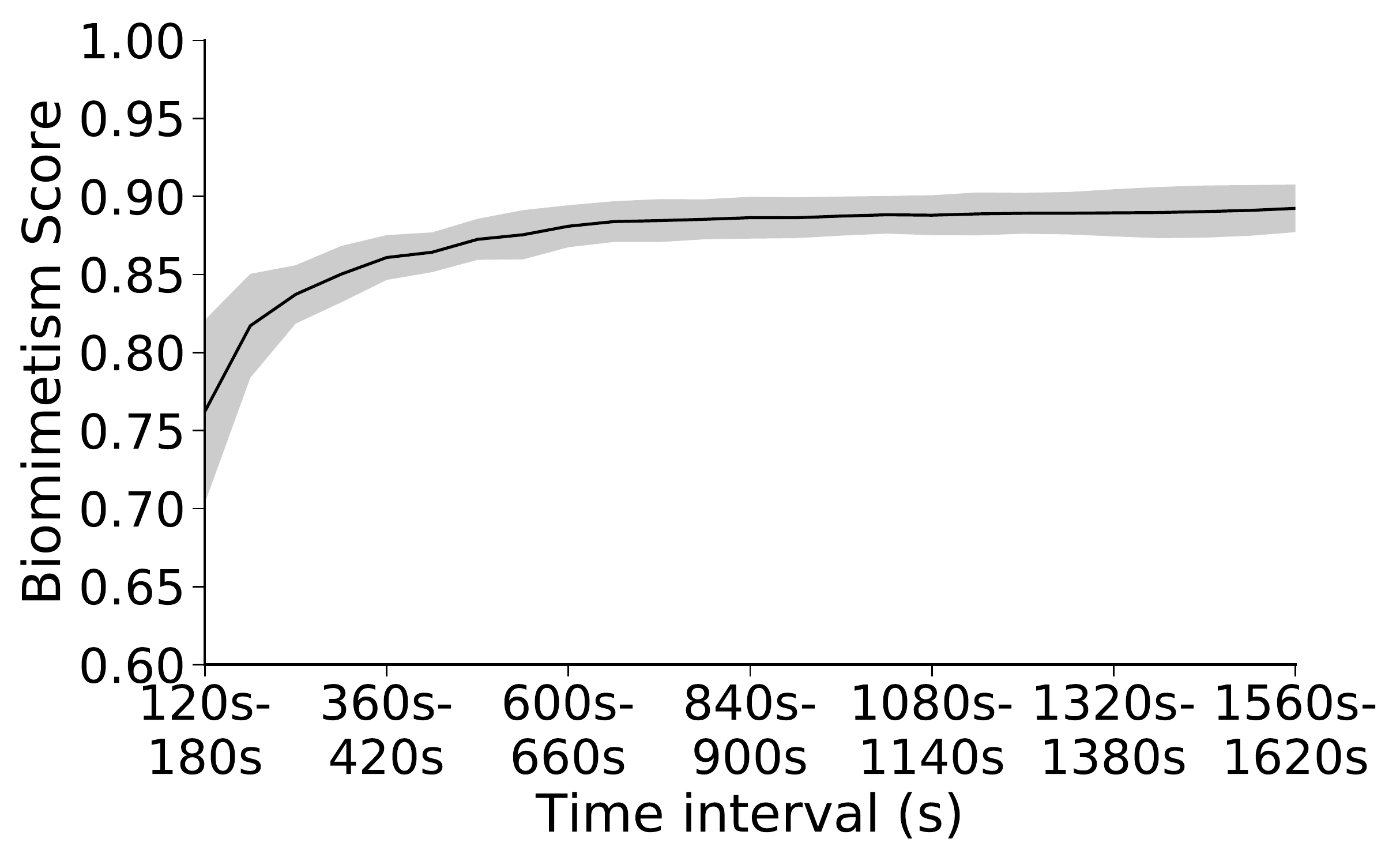}
\end{subfigure}

\begin{subfigure}[b]{0.45\textwidth}
\includegraphics[width=0.99\textwidth]{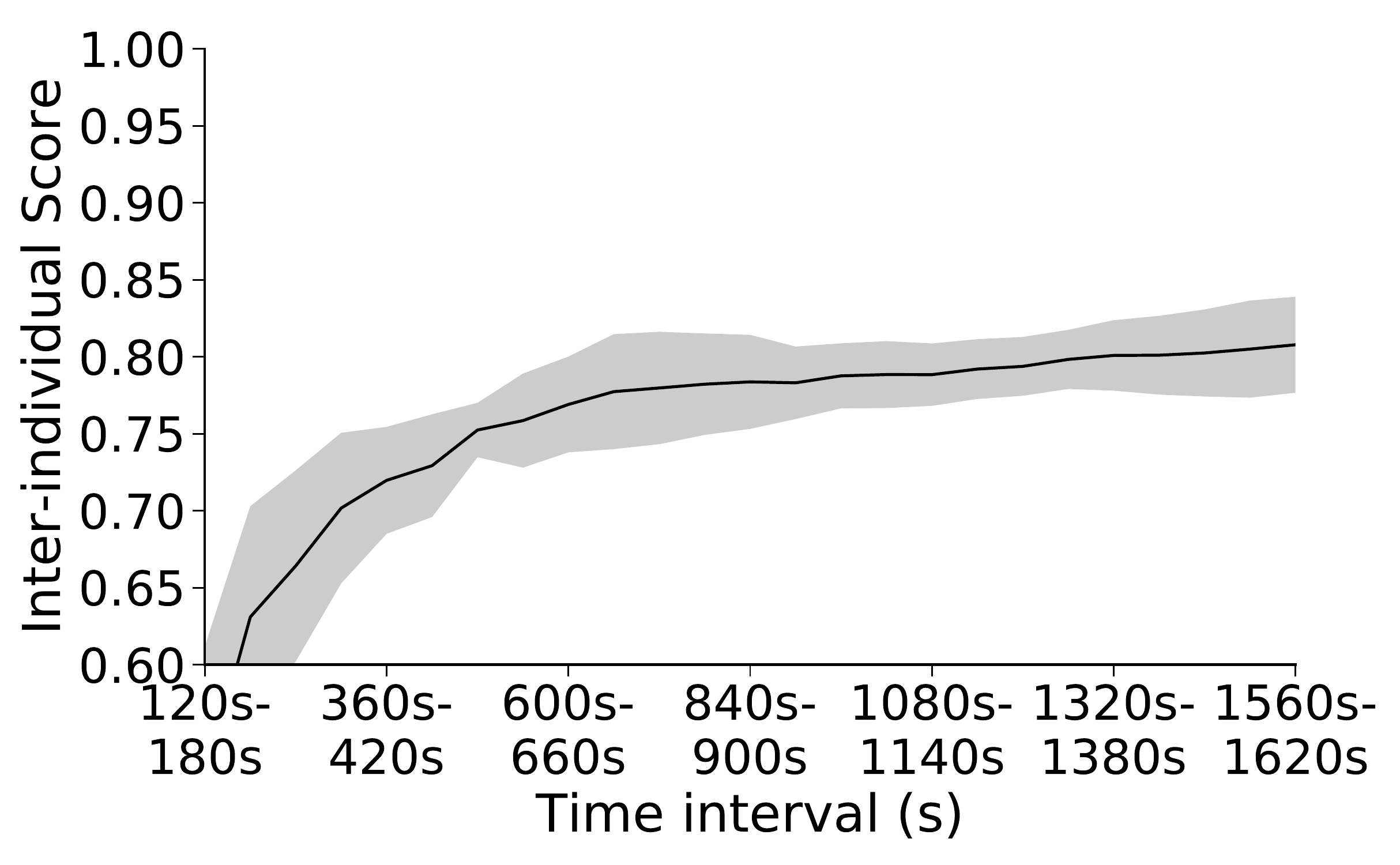}
\end{subfigure}
\begin{subfigure}[b]{0.45\textwidth}
\includegraphics[width=0.99\textwidth]{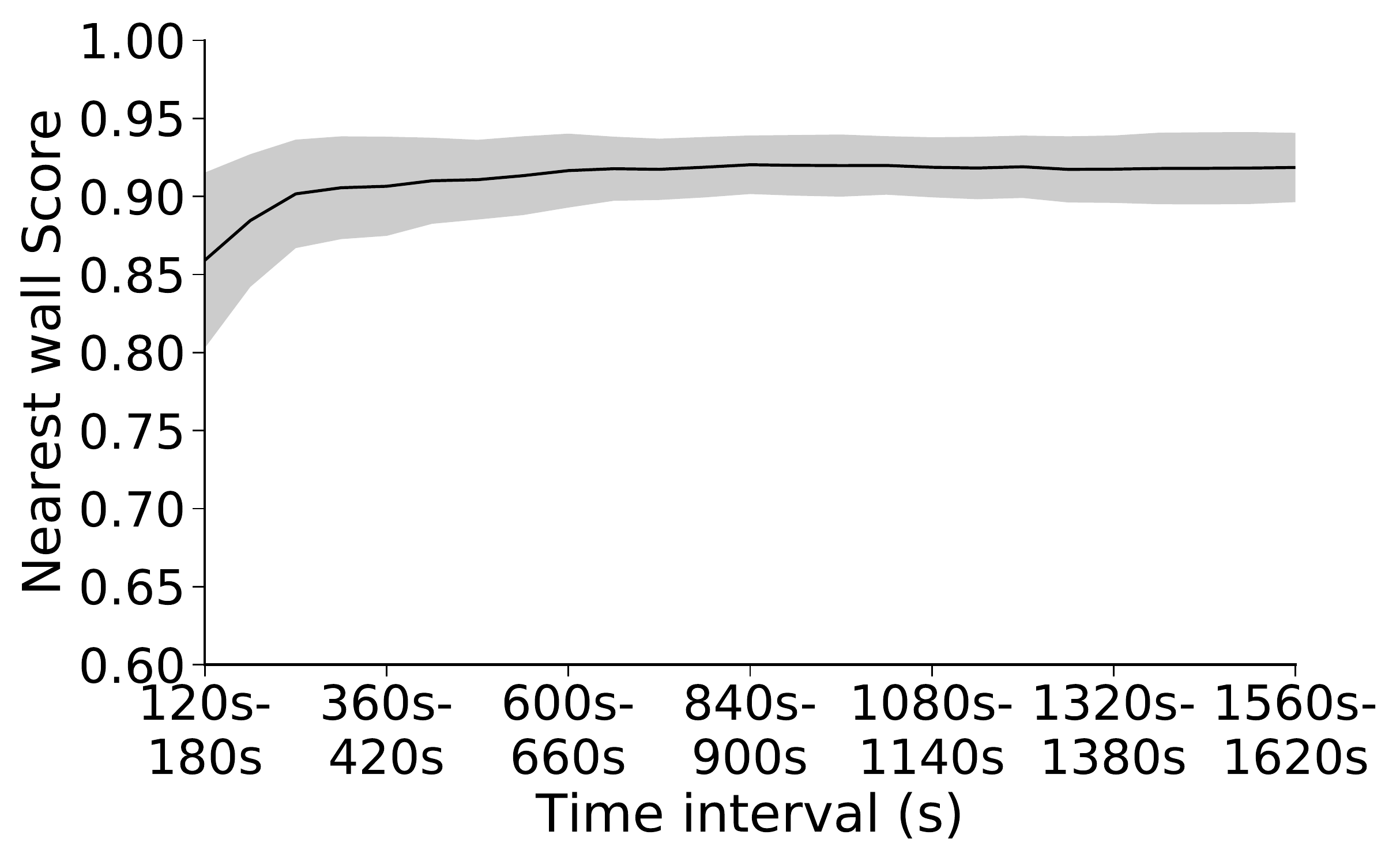}
\end{subfigure}

\begin{subfigure}[b]{0.45\textwidth}
\includegraphics[width=0.99\textwidth]{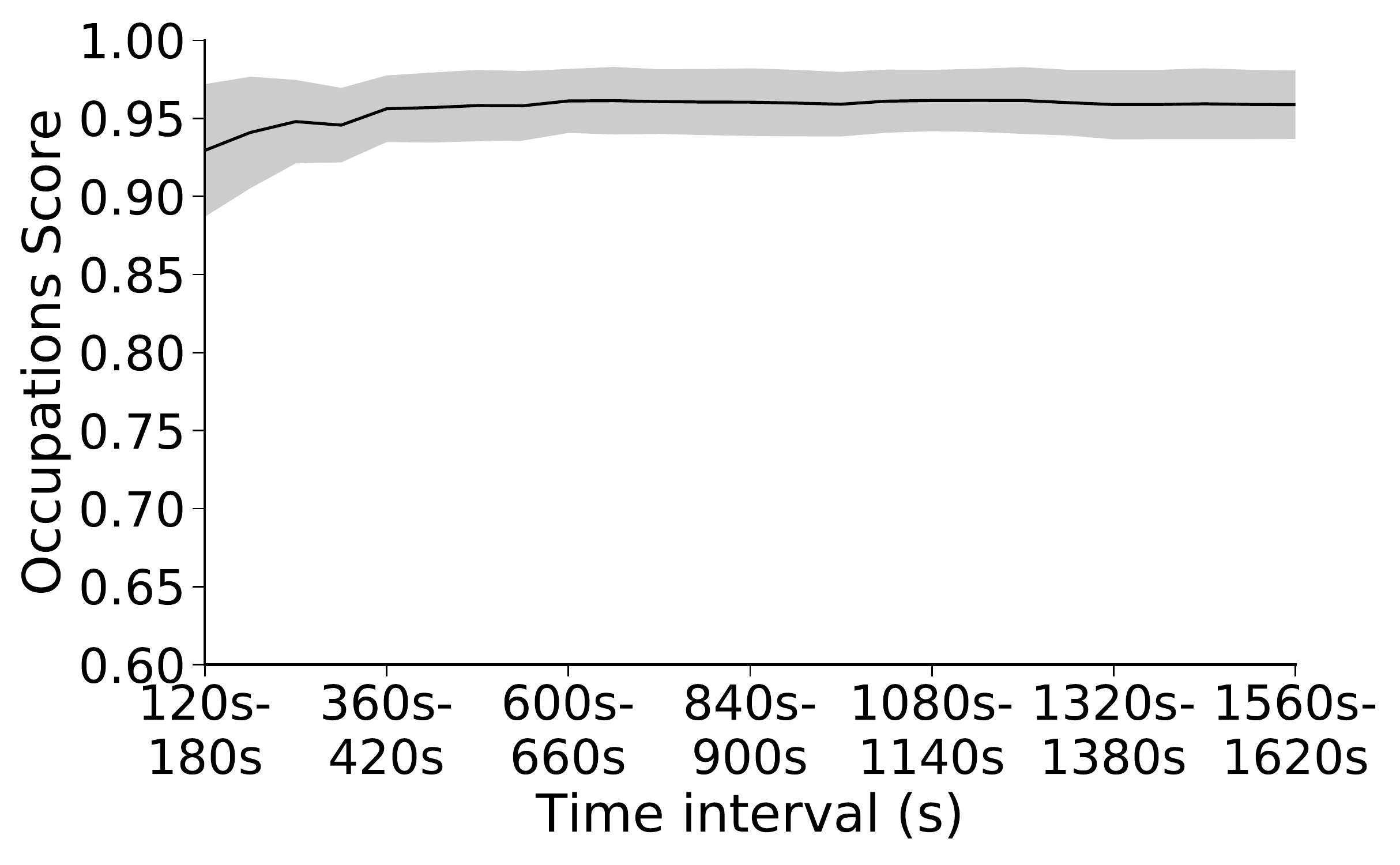}
\end{subfigure}
\begin{subfigure}[b]{0.45\textwidth}
\includegraphics[width=0.99\textwidth]{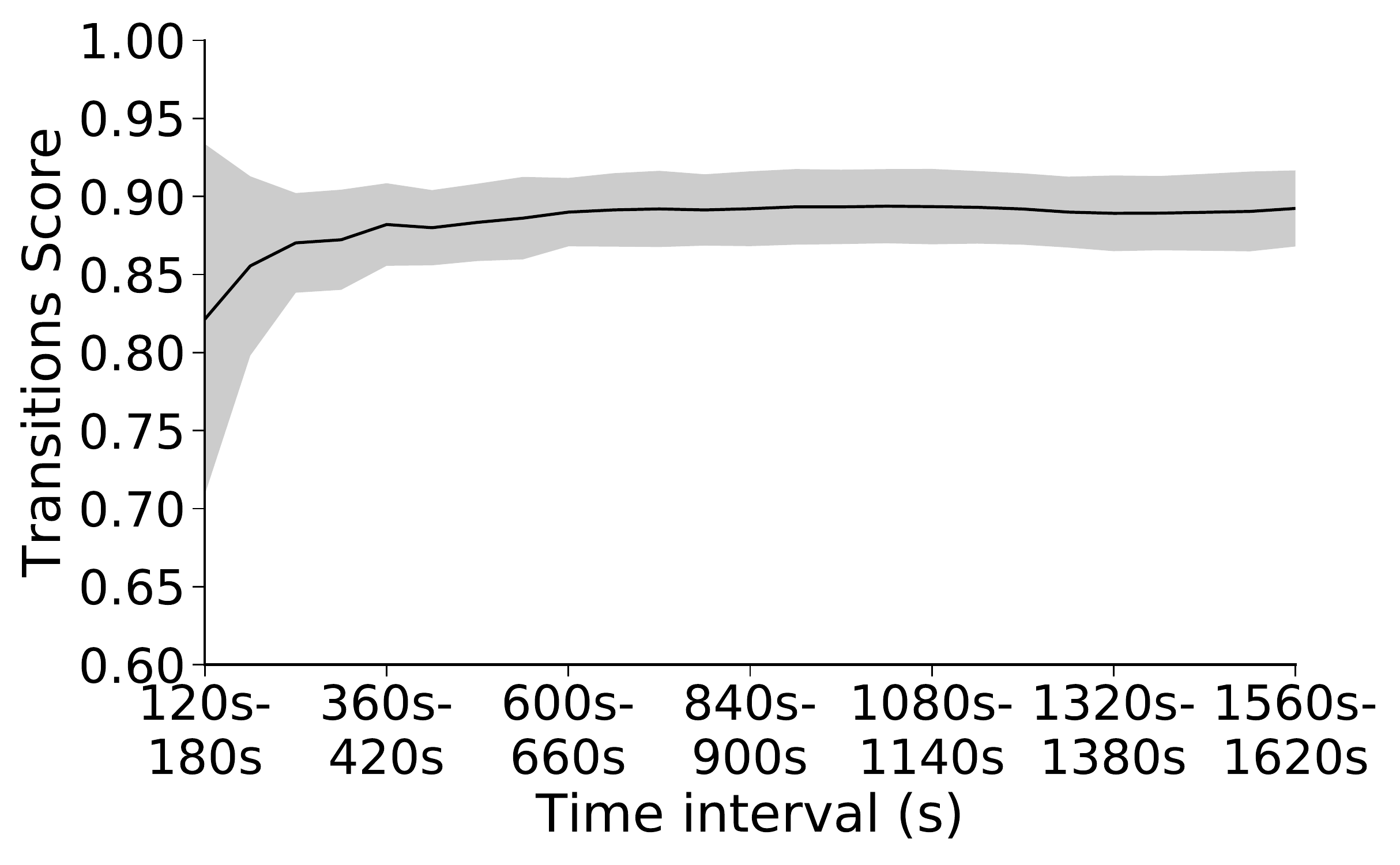}
\end{subfigure}

\caption{Similarity scores between the behaviour of the experimental fish and the behaviour of the best-performing individuals of the calibrated model at different time intervals of 10 different experiments. These scores are computed using data over $120s$, and starting from the third time interval ($120s$ to $180s$) to ensure gathering enough experimental data. In these plots, lines correspond to the mean scores across the 10 experiments, and the grey translucent areas correspond to the standard deviation. We consider four behavioural features to characterise the behaviour exhibited in each time interval. \textbf{Inter-individual distances} corresponds to the similarity in distribution of inter-individual distances between all agents in a specific zone and measures the capabilities of the agents to aggregate. \textbf{Distances to nearest wall} corresponds to the similarity in distribution of agent distance to their nearest wall, and assess their capability to follow the walls. \textbf{Occupations} corresponds to the similarity in probability of presence of the agent in each zone. \textbf{Transitions} corresponds to the similarity in probabilities of an agent to transition from one zone to another. The \textbf{Biomimetic score} corresponds to the geometric mean of the other scores.}
\label{fig:plotScores}
\end{center}
\end{figure}
\clearpage
}

We assessed the evolution of the similarity scores (defined in Sec.~\ref{sec:dataAnalysis}) between robot behaviour and fish behaviour, across sliding windows of $120s$ intervals of a set of 10 trials each one lasting $30$ minutes, starting in each trial from the second time interval ($120s$ to $180s$) to gather enough experimental data. These scores are compiled in Fig.~\ref{fig:plotScores}. The variance for the 10 trials is plotted as a grey area around the curves and remains rather small.

From its initial value of about $0.610$ (second time interval: $120s$ to $180s$), the average fitness (mean scores) fastly converges to values around $0.850$ starting from the fourth time interval ($0.827$ on $240s$ to $300s$). This is also observed for similarity scores of transitions and of distances to nearest wall. This shows that both of these behavioural features can be effectively optimised through an online evolutionary algorithm process, and remain stables during the experiment.  

The similarity score of zone occupation is particularly high at the beginning of the experiment, and is only slightly improved by our calibration methodology; this would suggests that room occupation is only slightly dependent of model parameters. This could be explained by a strong effect of room geometry (room size, and the general room configuration of the arena) over room occupation: rooms cover a larger area than the corridor. This could also be an effect of the aggregative behaviour exhibited by fish and by the model: the robot would follow the fish, which would tend to follow walls, thus explaining the relative invariance of the occupation score with respect to parameter values.

The variations in similarity score of inter-individual distances suggests changes of fish aggregative behaviour during the experiment. This could be explained by the fact that, while zebrafish tend to remain cohesive most of the time, they have the tendency of forming short-lived (a few seconds to a few minutes) sub-groups, especially when confronted to a fragmented environment~\cite{cazenille2017acceptation}.

\section{Discussion and Conclusion}
Animal-robot interaction studies employ simple robot behavioural model that are not adaptive or updated during the experiments. Often they are not biomimetic and do not close the interaction loop between the animals and the robots~\cite{cazenille2017acceptation}. Here we present a methodology to calibrate in real-time a multi-level context-dependent biomimetic model of fish behaviour to drive the behaviour of a biomimetic robot into a group of zebrafish. The model parameters are continuously refined to accurately correspond to the collective dynamics exhibited by fish during the experiments. The real-time nature of this calibration process allows the robot to react to changes in observed fish dynamics and cope with uncertainties.

Animals can present significant inter-individual behavioural differences. They can present significant differences in terms of personalities typically bold and shy types~\cite{toms2014back}. In most of the experiments individuals are selected randomly from a stock. Consequently, each group trial present differences depending on the characteristics of the individuals. Currently, the models are calibrated on a set of averaged experimental data and are not optimised to take into account inter-individuals differences. We present here a method to adapt in real-time the models and that is thus capable to cope with this issue. This method can reduce significantly the number of experimental trials necessary to calibrate the model.

Our approach is computationally intensive and use three networked computers to handle in real-time the tracking, the robot control, the data analysis and the model calibration tasks.

Our methodology builds on the work presented in~\cite{cazenille2017automated} by adding real-time capabilities to the calibration process. However, it also suffers from the same limitations. Namely, the model we calibrate must still be structurally defined empirically (\ie defining behavioural attractors, zones of the environment, etc) with ethological \textit{a-priori} knowledge about fish dynamics. The calibration process could also still be improved by taking into account additional behavioural metrics in the computation of similarity scores, either in term of collective dynamics (\eg agent groups aspects, residence time in a zone), individual behaviours (\eg agent trajectory aspects, curvature of trajectories). This could possibly be bypassed through the use of a calibration process without explicit similarity measure (e.g. GAN~\cite{goodfellow2014generative} or Turing Learning~\cite{li2016turing}).
Our behavioural model could be revised to account for collective departures of agents from one room to the other, as described in biological studies~\cite{collignon2017collective}. 

Additionally, our methodology could make use of global optimisation techniques designed to minimise the number of evaluations before reaching convergence, like Bayesian Optimisation~\cite{snoek2012practical,cully2015robots}. This would reduce calibration computation costs, and possibly reduce the time needed to accurately calibrate the models.

\section*{Acknowledgement}
{\small
This work was funded by EU-ICT project 'ASSISIbf', no 601074.}

\FloatBarrier

\end{document}